\documentclass[11pt]{article}

\usepackage[final]{acl}

\usepackage{times}
\usepackage{latexsym}

\usepackage[T1]{fontenc}

\usepackage[utf8]{inputenc}

\usepackage{microtype}

\usepackage{inconsolata}
\usepackage[most]{tcolorbox}

\usepackage{graphicx}
\usepackage{amsmath}
\usepackage{enumitem}
\usepackage{booktabs}
\usepackage{multirow}
\usepackage{algorithm}
\usepackage{algpseudocode}
\usepackage{graphicx}
\usepackage{subcaption}
\newcommand{\annotate}[3]{%
    #1\raisebox{-0.5ex}{\scriptsize\textcolor{#2}{#3}}%
}

\newtcolorbox{promptbox}[1][]{
  colback=gray!5!white,
  colframe=black!75!black,
  fonttitle=\bfseries,
  colbacktitle=gray!85!black,
  title=#1,
  width=\textwidth,           
  boxrule=0.3pt,
  arc=2pt,
  left=5pt, right=5pt, top=4pt, bottom=4pt,
  before skip=8pt, after skip=8pt,
  breakable,
  enhanced jigsaw
}

\title{When Is Thinking Enough? Early Exit via Sufficiency Assessment for Efficient Reasoning}

\author{
  Yang Xiang$^{1}$, 
  Yixin Ji$^{1}$, 
  Ruotao Xu$^{1}$,
  Dan Qiao$^{2}$,
  Zheming Yang$^{2}$,
  Juntao Li$^{1}$\thanks{\; Corresponding author.}, 
  \textbf{Min Zhang}$^{1}$ \\
  $^{1}$Soochow University \quad
  $^{2}$ByteDance\\
  \texttt{\{baldwin021129, jiyixin169\}@gmail.com} \quad
  \texttt{\{ljt, minzhang\}@suda.edu.cn}
}

\begin{document}
\maketitle
\begin{abstract}
Large reasoning models (LRMs) have achieved remarkable performance in complex reasoning tasks, driven by their powerful inference-time scaling capability.
However, LRMs often suffer from overthinking, which results in substantial computational redundancy and significantly reduces efficiency.
Early-exit methods aim to mitigate this issue by terminating reasoning once sufficient evidence has been generated, yet existing approaches mostly rely on handcrafted or empirical indicators that are unreliable and impractical.
In this work, we introduce \textbf{Dynamic Thought Sufficiency in Reasoning (DTSR)}, a novel framework for efficient reasoning that enables the model to dynamically assess the sufficiency of its chain-of-thought (CoT) and determine the optimal point for early exit.
Inspired by human metacognition, DTSR operates in two stages: (1) \textit{Reflection Signal Monitoring}, which identifies reflection signals as potential cues for early exit, and (2) \textit{Thought Sufficiency Check}, which evaluates whether the current CoT is sufficient to derive the final answer.
Experimental results on the Qwen3 models show that DTSR reduces reasoning length by 28.9\%–34.9\% with minimal performance loss, effectively mitigating overthinking.
We further discuss overconfidence in LRMs and self-evaluation paradigms, providing valuable insights for early-exit reasoning.
\end{abstract}
\section{Introduction}
The success of large reasoning models (LRMs), such as OpenAI’s o1/o4~\cite{openai2024openaio1card,openai2025o4}, Gemini 2.5 Pro~\cite{comanici2025gemini25pushingfrontier}, DeepSeek-R1~\cite{Guo2025DeepSeek}, and Qwen3~\cite{yang2025qwen3technicalreport}, represents a significant breakthrough in the evolution of LLMs.
LRMs engage in long chain-of-thought (CoT) reasoning before producing answers.
The inference-time scaling capability~\cite{DBLP:journals/corr/abs-2503-09567,ji2025surveytesttimecomputeintuitive} allows them to achieve remarkable performance in challenging reasoning tasks (e.g. mathematics, programming, and scientific question answering), and show human-like reasoning patterns, such as problem decomposition, verification and backtracking.

Nevertheless, recent studies~\citep{chen2025do,cuadron2025dangeroverthinkingexaminingreasoningaction} have shown that LRMs often suffer from overthinking: even on simple problems, they tend to consume excessive tokens by repeatedly verifying answers or exploring alternative strategies after reaching the correct solution.
Such redundant reasoning is unnecessary and leads to inefficiency.
To address this, several prior works have proposed early-exit strategies that stop the reasoning process once the LRM has generated sufficient evidence to solve the problem.
The main challenge of early-exit strategies is to determine the optimal exit point, where LRMs achieve near-lossless performance with minimal reasoning cost.
Dynasor-CoT~\citep{fu2025reasoning} first employs the consistency of intermediate answers to determine when to exit reasoning; however, it still generates additional token intervals after producing the correct answer to verify its correctness, thereby missing the optimal exit point.
DEER~\citep{yang2025dynamicearlyexitreasoning} and CGRS~\citep{huang2025efficientreasoninglargereasoning} 
employ the entropy of the intermediate result as a confidence measure to determine whether LRMs reach the optimal exit point.
Nevertheless, using answer confidence as a criterion for reasoning exit is not always reliable, as reasoning models can exhibit overconfidence even when their answers are incorrect~\citep{mei2025reasoninguncertaintyreasoningmodels}.
Moreover, such methods that rely on probing intermediate answers are limited to tasks with short, fixed-format answers.
For tasks involving long answers or open-ended generation without a definitive answer, intermediate answers are difficult to extract, rendering these methods impractical.

To overcome these limitations, this work aims to establish a reliable and generalizable criterion for early-exit.
Unlike LLM reasoning, human cognition involves metacognitive~\citep{Flavell1979MetacognitionAC,fleur2021metacognition} abilities, which allow people to monitor, evaluate, and regulate their thoughts, and to stop thinking once the process is deemed sufficient.
Inspired by this, we wonder whether LRMs can similarly determine the optimal point to exit reasoning by continuously assessing the sufficiency of their current thoughts?
To answer this, we propose Dynamic Thought Sufficiency in Reasoning (DTSR), a novel framework that enables models to dynamically evaluate whether their current CoT is sufficient to solve the problem.
DTSR consists of two stages: 1) Reflection Signal Monitoring: LRMs often exhibit reflective behaviors after producing an answer, typically triggered by specific reflection signals.
We identify these signals as potential exit points for reasoning. 2) Thought Sufficiency Check: once a reflection signal is detected, the model evaluates the sufficiency of its current reasoning. If the sufficiency score is high, reasoning stops and the solution is output; otherwise, the model continues along the reasoning path until the next reflection signal occurs.

To evaluate the effectiveness of our proposed method, we conducted extensive experiments across multiple LRMs.
The results show that DTSR significantly outperforms baseline approaches, achieving a favorable balance between accuracy and efficiency.
This also demonstrates that reasoning models can reliably assess the sufficiency of their thought, providing valuable insights and guidance for advancing efficient reasoning.

Overall, our contributions are as follows:
\begin{itemize} [leftmargin=*]
\setlength{\itemsep}{0pt}
\setlength{\parskip}{0pt}
    \item We propose a novel framework for efficient reasoning that enables the model to dynamically assess the sufficiency of its CoT and determine the optimal point for early exit.

    \item Extensive experiments conducted on Qwen3 series models of various sizes demonstrate that DTSR consistently outperforms baseline methods, achieving a 28.9\%–34.9\% reduction in generation length with almost no loss in performance, thereby significantly mitigating overthinking.

    \item Further discussion of overconfidence in LRMs and the self-evaluation paradigm offers a fresh perspective and valuable insights for guiding early-exit reasoning.

\end{itemize}

\section{Methodology}

\begin{figure*}[t]
	\centering
    \includegraphics[width=\textwidth]{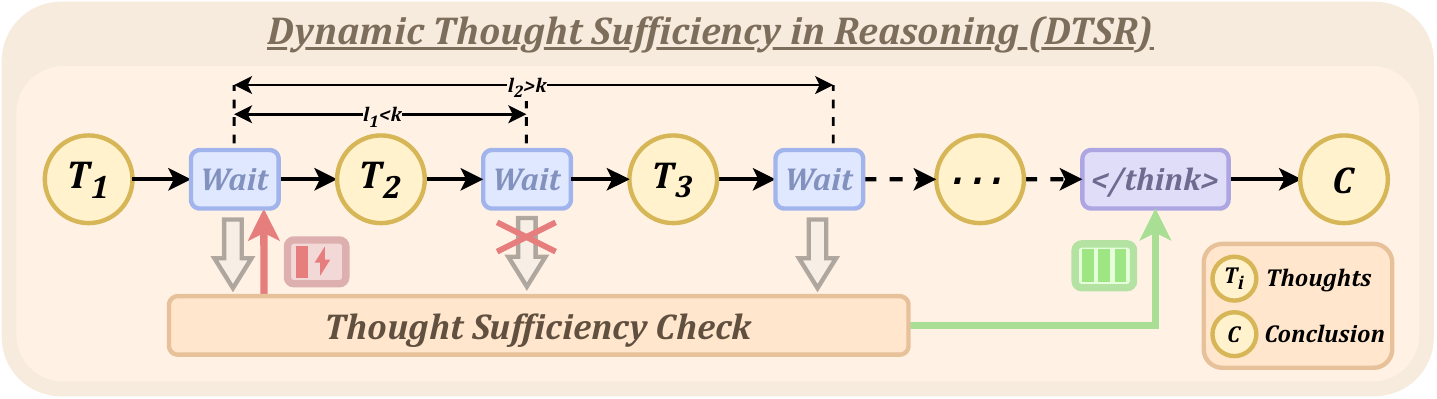}
    \caption{Overview of the Dynamic Thought Sufficiency in Reasoning (DTSR) method. Upon a reflection signal, the model evaluates the sufficiency of its previous thought. If the thought is insufficient, reasoning continues; checks are skipped within k tokens. Once the thought is sufficient, \texttt{</think>} is appended and a conclusion is generated.}
  \label{fig:method}
\end{figure*}

\subsection{Preliminaries}

\paragraph{Generation Patterns of Reasoning Models}
The responses of LRMs are typically structured in two stages: a reasoning phase, in which the model engages in slow, human-like thinking, encapsulated within the \texttt{<think>} and \texttt{</think>} delimiters, followed by a solution phase that summarizes the reasoning and produces the final answer.
During the thinking process, the model exhibits complex reasoning behaviors, such as problem decomposition, strategy exploration, output verification, and self-correction~\citep{gandhi2025cognitive,bogdan2025thoughtanchorsllmreasoning}.
To trigger these reflective behaviors, LRMs often generate tokens like ``Wait'', ``Alternatively'', or ``Let me check'' as cues.
However, once the correct answer has been reached, further verification becomes redundant, leading to unnecessary reasoning cost~\citep{chen2025do}.

\paragraph{Early Exit in Reasoning}
Early exit is a promising strategy for mitigating redundant reasoning.
~\citet{muennighoff2025s} first introduces a simple budget forcing method: when generation reaches the maximum token budget, the model is forced to append the end-of-thinking token and “Final Answer:” to terminate reasoning and output its current answer.
Dynasor-CoT~\citep{fu2025reasoning} probes intermediate answers via designed prompts at fixed intervals and terminates reasoning once consecutive answers are consistent.
It partially mitigates overthinking but still generates extra token intervals to verify correctness after the correct answer is produced, failing to achieve optimal exit point.
In contrast, DEER~\citep{yang2025dynamicearlyexitreasoning} 
outputs intermediate answers at reasoning transition points (e.g., “Wait”) and uses entropy to estimate their confidence, terminating the reasoning process when achieving high confidence.

However, previous methods mainly rely on handcrafted or statistical criteria, which are unreliable for determining whether the current reasoning process is sufficient to yield the correct answer.
~\citet{mei2025reasoninguncertaintyreasoningmodels} find that even state-of-the-art reasoning models are overconfident, particularly for incorrect answers, and this overconfidence grows with the longer reasoning process.
Consequently, a substantial gap exists between token probabilities and the actual correctness of answers.
From a cognitive perspective, humans rarely produce explicit intermediate answers to decide when to stop thinking.
Instead, through metacognitive processes~\citep{fleur2021metacognition}, they internally assess whether their thoughts are sufficient to reach a final answer.
Inspired by this, enabling reasoning models to self-assess the sufficiency of their thought to determine if it can yield a final answer is highly intuitive.
Therefore, we propose Dynamic Thought Sufficiency in Reasoning (DTSR), a method that enables models to assess whether their current thought is sufficient to derive the final answer.

\subsection{Our Proposal: Dynamic Thought Sufficiency in Reasoning}
As shown in Figure~\ref{fig:method}, we present DTSR, a method that enables the model to dynamically evaluate the sufficiency of its reasoning process.
Our method consists of two stages: reflection signal monitoring and thought sufficiency check.

\paragraph{Reflection Signal Monitoring}
Our objective is to exit the reasoning process at the optimal point. Since assessing thought sufficiency after every generated token is computationally impractical, we instead seek to identify potential exit signals that indicate when verification should be triggered.
To this end, we conduct an empirical analysis of reasoning trajectories generated by Qwen3-32B. For each trajectory, we identify its optimal early-exit point, defined as the earliest position in the reasoning process at which the model is already capable of producing a correct answer (Figure~\ref{fig:example}).
We find that these optimal exit points are often followed by explicit self-reflection or answer-verification behaviors, manifested through cues such as “Wait”, “Alternatively”, and phrases beginning with “But” (e.g., “But wait”, “But let me check”).
Based on these observations, we construct a reflection signal set, the complete list of which is shown in Table~\ref{tab:Reflection_Signal}.

\paragraph{Thought Sufficiency Check}

The LRM continues reasoning until a reflection signal is encountered. Upon detecting such a signal, the model performs a thought sufficiency check on the reasoning trace generated so far to determine whether it can terminate further thinking and produce a final solution. Specifically, at each checkpoint, the original question together with the current chain-of-thought (CoT) is fed into a dedicated thought sufficiency check prompt template (shown in Figure~\ref{fig:template}). The model is then prompted to generate a scalar score indicating the sufficiency of the current reasoning. A score exceeding the threshold $\tau$ indicates that the model deems its reasoning sufficient; in this case, we append a termination marker \texttt{</think>} to halt further reasoning and allow the model to output the final answer. Conversely, if the score falls below the threshold $\tau$, the reasoning is considered insufficient, and the model resumes generation from the last reflection signal, continuing the reasoning process until the next checkpoint. This procedure is repeated until the reasoning process is complete.
Unlike prior approaches that rely on handcrafted heuristics or empirical external metrics, our method determines the reasoning exit point through the model’s internal self-evaluation.
In this paradigm, the model assesses the sufficiency of its previously generated reasoning from a third-person perspective, treating the reasoning trace as an external object rather than directly evaluating itself.
Further discussion of this can be found in Section \ref{selfeval}.

Moreover, reflection signals frequently occur in close succession, causing repeated sufficiency checks to yield similar outcomes and resulting in redundant computation.
To mitigate this issue, we enforce a minimum token interval $k$ between consecutive sufficiency checks.
Concretely, a reflection signal triggers a thought sufficiency check only if at least $k$ tokens have been generated since the previous check; otherwise, the current signal is skipped and the process continues (Algorithm~\ref{alg:dtsr}).

\begin{algorithm}[t]
\caption{Dynamic Thought Sufficiency in Reasoning (DTSR)}
\label{alg:dtsr}
\begin{algorithmic}[1]
\State \textbf{Init:} model $\mathcal{M}$, question $Q$, prompt $q\_cot$, reflection signals $\mathcal{S}$, min interval $k$, threshold $\tau$, max length $max\_len$
\State $L \gets q\_cot + Q$, reasoning trace $T \gets \emptyset$, counter $c \gets 0$
\While{len($L$) $< max\_len$}
    \State Generate token $t$, $T \gets T + t$, $c \gets c + 1$
    \If{$t \in \mathcal{S}$ \textbf{and} $c \ge k$}
        \State Build sufficiency prompt with $(Q,T)$
        \State Obtain score $s \in [0,100]$ from $\mathcal{M}$

        \If{$s \ge \tau$}
            \State $T \gets T + \texttt{</think>}$ 
        \Else
            \State Continue reasoning, $c \gets 0$
        \EndIf
    \EndIf
\EndWhile
\State \Return $T$
\end{algorithmic}
\end{algorithm}

\section{Experiments}

\begin{table*}[!t]
\centering
\small
\label{table1}
\renewcommand{\arraystretch}{1}
\begin{tabular}{@{}lccccccccccll@{}} 
\toprule
 \multirow{2}{*}{\textbf{Method}} 
 & \multicolumn{2}{c}{\textbf{GSM8K}} & \multicolumn{2}{c}{\textbf{MATH-500}} & \multicolumn{2}{c}{\textbf{AMC}} & \multicolumn{2}{c}{\textbf{GPQA}} & \multicolumn{2}{c}{\textbf{Olympiad}} & \multicolumn{2}{|c}{\textbf{Overall}} \\
    & {Acc$\uparrow$} & {Tok$\downarrow$} & {Acc$\uparrow$} & {Tok$\downarrow$} & {Acc$\uparrow$} & {Tok$\downarrow$} & {Acc$\uparrow$} & {Tok$\downarrow$} & {Acc$\uparrow$} & {Tok$\downarrow$} & {Acc$\uparrow$} & {Tok$\downarrow$} \\ 
\midrule
\multicolumn{13}{c}{\textit{\textbf{Qwen3-8B}}} \\
\midrule
\textit{Vanilla} & 96.0 & 2208 & 94.4 & 5067 & 92.5 & 7232 & 60.1 & 8861 & 66.4 & 9182 & \multicolumn{1}{|l}{81.9} & 6510 \\
\textit{NoThinking} & 92.6 & 303 & 86.0 & 1067 & 72.5 & 2264 & 53.0 & 1271 & 48.3 & 2385 & \multicolumn{1}{|l}{70.5} & 1458 \\
\textit{NoWAIT} & 94.9 & 1307 & 92.8 & 3485 & 90.0 & 5448 & 55.6 & 6920 & 62.4 & 6998 & \multicolumn{1}{|l}{79.1} & 4832 \\
\textit{DEER} & 95.1 & 963 & 93.4 & 2549 & 87.5 & 5155 & 57.1 & 8306 & 63.4 & 5768 & \multicolumn{1}{|l}{79.3} & 4532 \\
\textit{DTSR (ours)} & 95.1 & 1152 & 94.0 & 2359 & 90.0 & 4507 & 59.6 & 7942 & 66.4 & 6180 & \multicolumn{1}{|l}{{\textbf{\annotate{81.0}{red}{-0.9}}}} & {\textbf{\annotate{4428}{blue}{-32.0\%}}} \\
\midrule
\multicolumn{13}{c}{\textit{\textbf{Qwen3-14B}}} \\
\midrule
\textit{Vanilla} & 96.2 & 1672 & 95.4 & 4503 & 97.5 & 6562 & 65.7 & 7513 & 67.0 & 8555 & \multicolumn{1}{|l}{84.4} & 5761\\
\textit{NoThinking} & 95.1 & 278 & 88.2 & 852 & 77.5 & 2065 & 55.6 & 1286 & 51.7 & 2029 & \multicolumn{1}{|l}{73.6} & 1302 \\
\textit{NoWAIT} & 95.8 & 1133 & 92.8 & 3307 & 97.5 & 5035 & 64.1 & 6775 & 65.6 & 6670 & \multicolumn{1}{|l}{83.2} & 4584 \\
\textit{DEER} & 95.3 & 896 & 94.4 & 2601 & 97.5 & 4514 & 62.6 & 7105 & 65.1 & 6717 & \multicolumn{1}{|l}{83.0} & 4367 \\
\textit{DTSR (ours)} & 96.2 & 849 & 95.0 & 2247 & 97.5 & 3975 & 66.2 & 5916 & 68.9 & 5755 & \multicolumn{1}{|l}{{\textbf{\annotate{84.8}{red}{+0.4}}}} & {\textbf{\annotate{3748}{blue}{-34.9\%}}} \\
\midrule
\multicolumn{13}{c}{\textit{\textbf{Qwen3-32B}}} \\
\midrule
\textit{Vanilla} & 96.6 & 1707 & 94.8 & 4489 & 95.0 & 6807 & 68.2 & 6906 & 69.0 & 8280 & \multicolumn{1}{|l}{84.7} & 5638\\
\textit{NoThinking} & 94.8 & 283 & 86.4 & 958 & 70.0 & 1738 & 54.6 & 561 & 52.4 & 2076 & \multicolumn{1}{|l}{71.6} & 1123 \\
\textit{NoWAIT} & 95.8 & 1212 & 93.4 & 3285 & 95.0 & 4894 & 66.7 & 5436 & 65.8 & 6598 & \multicolumn{1}{|l}{83.3} & 4285 \\
\textit{DEER} & 95.5 & 979 & 94.0 & 2881 & 92.5 & 4966 & 67.2 & 6395 & 66.4 & 6370 & \multicolumn{1}{|l}{83.1} & 4318 \\
\textit{DTSR (ours)} & 96.0 & 849 & 95.0 & 2660 & 95.0 & 4617 & 69.2 & 5677 & 67.7 & 6249 & \multicolumn{1}{|l}{{\textbf{\annotate{84.6}{red}{-0.1}}}} & {\textbf{\annotate{4010}{blue}{-28.9\%}}} \\
\bottomrule
\end{tabular}
\caption{Experimental results across various reasoning models. "Acc" denotes accuracy, and "Tok" denotes token count. $\uparrow$ indicates that higher values are better, while $\downarrow$ indicates that lower values are better.}
\label{table:main}
\end{table*}

\subsection{Experimental Setup}

\paragraph{Benchmarks and Metrics}

To comprehensively evaluate the reasoning capabilities of LRMs, we select six widely used benchmarks: GSM8K~\cite{cobbe2021trainingverifierssolvemath}, MATH-500~\cite{hendrycks2021measuring}, AMC 2023, OlympiadBench~\cite{he-etal-2024-olympiadbench}, GPQA Diamond~\cite{rein2024gpqa} and LiveCodeBench~\cite{jain2025livecodebench}.
Our goal is to significantly reduce the number of tokens generated during the reasoning process while maintaining high reasoning performance.
To this end, we employ two primary metrics: Accuracy (Acc) and Token Count (Tok).
Acc measures the proportion of correct final answers, calculated as the average pass@1 score, whereas Tok represents the average number of tokens generated per sample, serving as an indicator of reasoning efficiency.

\paragraph{Models}
We conduct extensive experiments on the Qwen3~\cite{yang2025qwen3technicalreport} series of models.
Our selection spans different model sizes (8B, 14B, 32B) of reasoning models to validate the generalizability of our method.
Despite their strong reasoning ability, these models often exhibit overthinking during the reasoning process.

\paragraph{Baselines}

We compare DTSR with the vanilla LRM as well as existing state-of-the-art training-free efficient reasoning methods, including NoThinking~\citep{ma2025reasoningmodelseffectivethinking}, NOWAIT~\citep{wang2025waitdontneedwait}, and DEER~\citep{yang2025dynamicearlyexitreasoning}.
Vanilla allows the model to generate answers directly without any intervention.
NoThinking prompts the model to skip the reasoning stage and answer the question directly.
NOWAIT suppresses explicit self-reflection by masking reflective tokens such as “wait” and “hmm” during the reasoning process.
DEER makes early-exit decisions based on the confidence scores of intermediate answers.
In addition, we compare DTSR with recent training-based methods~\citep{arora2025training,dai2025sgrpo}.

\paragraph{Implementation Details}
All experiments are conducted using the vLLM framework to ensure high throughput and memory-efficient inference.
For the decoding strategy, we set the temperature to 0.6 and top-$p$ to 0.95.
The maximum generation length is set to 16k tokens, and each dataset is generated three times, with the final results reported as the average pass@1.
In our method, the thought sufficiency threshold $\tau$ is set to 100, and the minimum token interval $k$ is set to 64.

\subsection{Experimental Results}
\paragraph{Overall Performance}
The results in Table~\ref{table:main} show that DTSR consistently outperforms the baseline methods across three model sizes and five benchmark datasets.
Compared with vanilla reasoning models, DTSR achieves almost the same level of accuracy without any degradation, while reducing sequence length by 28.9\%–34.9\%, leading to more efficient inference.
Notably, when applied to Qwen3-14B, DTSR even improves performance on challenging reasoning tasks such as GPQA and OlympiadBench, underscoring its effectiveness.

\paragraph{Comparison with SOTAs}

We compare DTSR with several recent training-free efficient reasoning methods.
As shown in Table~\ref{table:main}, DTSR effectively shortens reasoning length without compromising accuracy, whereas other baselines struggle to balance efficiency with performance.
For instance, NoThinking enforces models to bypass the reasoning process through prompting, which indeed shortens sequences but deviates from the natural generation paradigm and severely limits reasoning capability.
NoWAIT mitigates redundant output by masking frequent reflective tokens, yet this restriction on the model’s inherent self-reflection results in significant performance degradation on complex reasoning tasks.
DEER relies on the confidence scores of intermediate answers, yet models often exhibit high confidence even in incorrect ones. Moreover, it focuses only on partial answers while neglecting the global reasoning process, making it unreliable.
In contrast, DTSR performs self-assessment over the entire reasoning process, which accounts for global coherence while mitigating overconfidence, thereby achieving superior overall performance.

\begin{figure}[t]
    \centering
    \includegraphics[width=\columnwidth]{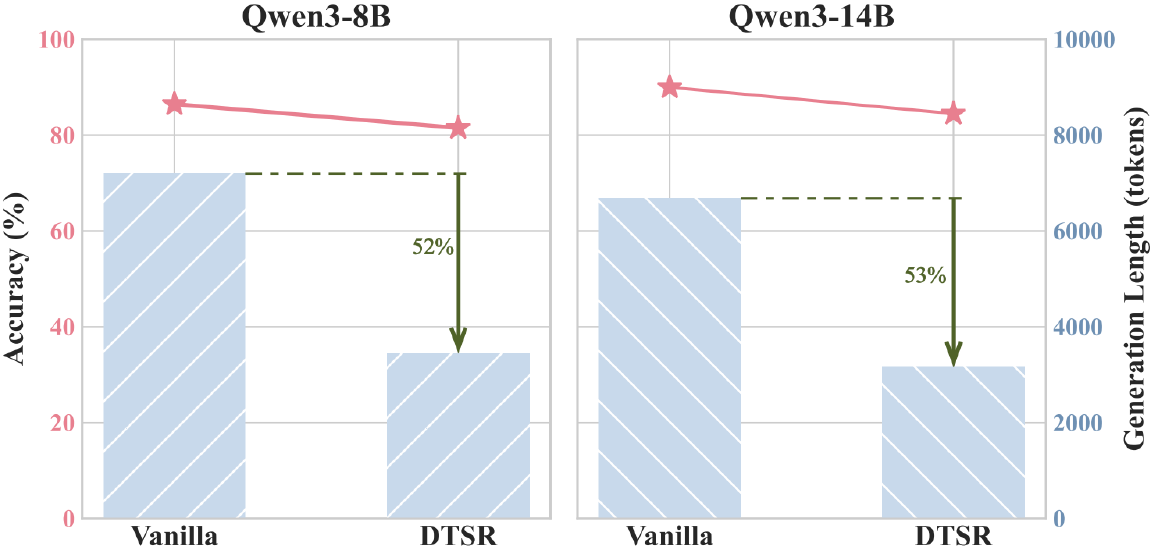}
    \caption{Performance of DTSR on LiveCodeBench.}
    \label{fig:programme}
\end{figure}

We also present a comparison between DTSR and training-based methods in Table~\ref{table:training_based}.
The results show that DTSR achieves a generation length comparable to training-based approaches. 
Although DTSR exhibits slightly lower accuracy due to the absence of additional training, these results demonstrate that DTSR remains competitive with training-based methods while requiring no extra training.

\begin{figure}[t]
    \centering
    \includegraphics[width=\columnwidth]{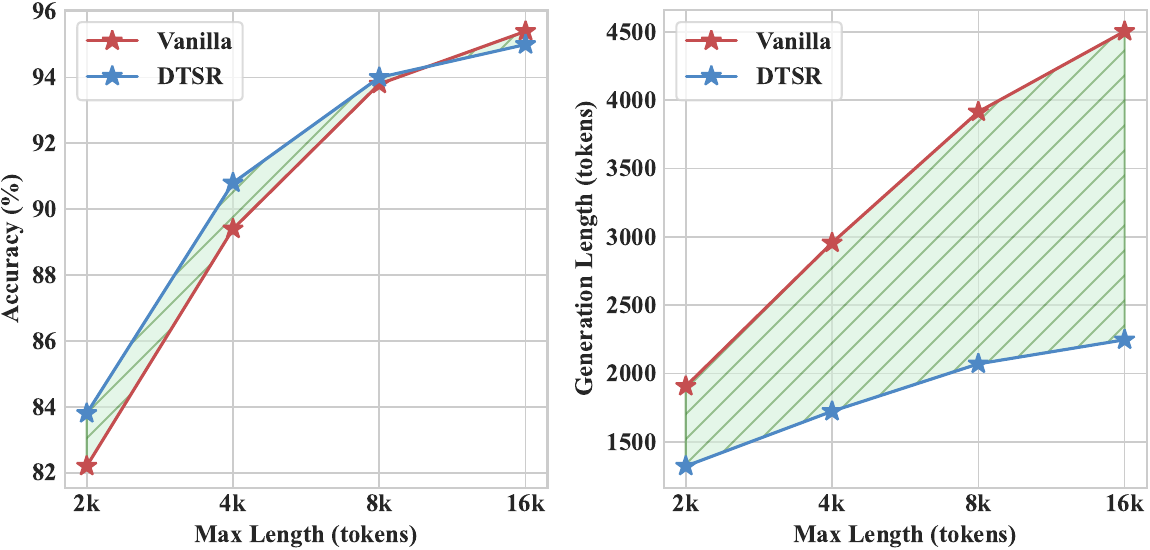}
    \caption{Comparison of Vanilla and DTSR on MATH-500 as the token budget increases.}
    \label{fig:token_budget}
\end{figure}

\paragraph{Performance on Programming Tasks}

To evaluate the generalization of DTSR across different domains, we additionally assess its performance on programming tasks.
Figure~\ref{fig:programme} presents the evaluation results of Qwen-8B and Qwen-14B on LiveCodeBench.
Compared to mathematical and scientific tasks, the reasoning process in programming tasks tends to be more redundant.
DTSR is able to reduce the generation length by over 50\% while incurring only minimal performance degradation. 
Overthinking is particularly pronounced in programming tasks, as they typically involve longer answers, requiring the model to spend more tokens on verification and reflection.
In many cases, the model is already capable of generating correct code during the intermediate reasoning steps, repeated verification prolongs this process.

\section{Analysis and Discussion}
\subsection{Ablation Study}

\paragraph{Performance with different token budget}

We expand the maximum generation token budget from 2k to 16k to compare Vanilla and DTSR in terms of accuracy and generation length.
Figure~\ref{fig:token_budget} shows that DTSR consistently generates shorter sequences across all tested budgets, with accuracy only showing a slight fluctuation at 16k, while outperforming Vanilla in the other settings.
Notably, the difference in sequence length between Vanilla and DTSR grows as the budget increases, suggesting that DTSR achieves greater token savings under larger budgets.
In practice, our method can dynamically adjust the token budget to suit diverse application scenarios, thereby improving flexibility and efficiency in real-world deployments.

\paragraph{Impact of token interval $k$}

\begin{figure}[t]
    \centering
    \includegraphics[width=\columnwidth]{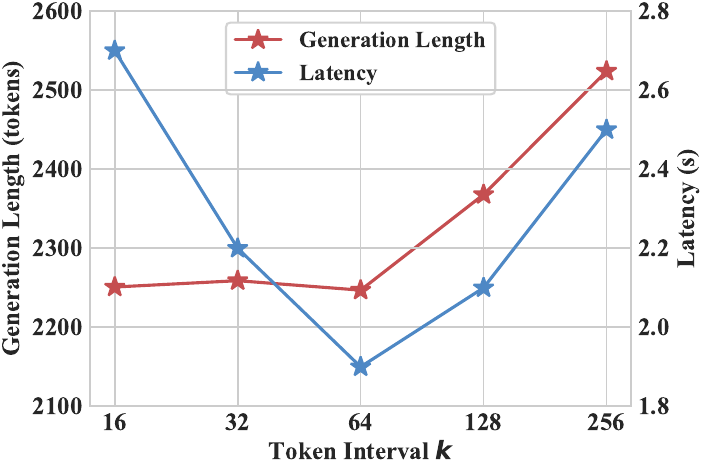}
    \caption{Impact of token interval $k$.}
    \label{fig:token_k}
\end{figure}

To begin with, varying the token interval $k$ has almost no effect on accuracy, since the criterion for assessing thought sufficiency remains unchanged.
Instead, $k$ primarily affects two factors: generation length and inference latency.
Figure~\ref{fig:token_k} presents the results of DTSR on MATH-500 under different token intervals.
When $k \leq 64$, the generation length remains nearly constant.
However, once $k > 64$, the length starts to increase as $k$ grows.
This occurs because overly large token intervals may cause the model to miss the optimal exit points for certain problems, thereby delaying exit.
Regarding latency, an overall U-shaped trend can be observed.
When $k \leq 64$, the reasoning latency increases due to the frequent thought-sufficiency checks required at smaller intervals. In contrast, when $k > 64$, although the number of checks is reduced, the model tends to generate longer reasoning traces after missing optimal exit points, which in turn leads to increased latency.
Consequently, $k = 64$ represents a balanced trade-off between generation length and latency, making it a desirable choice in practice.

\paragraph{Impact of threshold $\tau$}

\begin{figure}[t]
    \centering
    \includegraphics[width=\columnwidth]{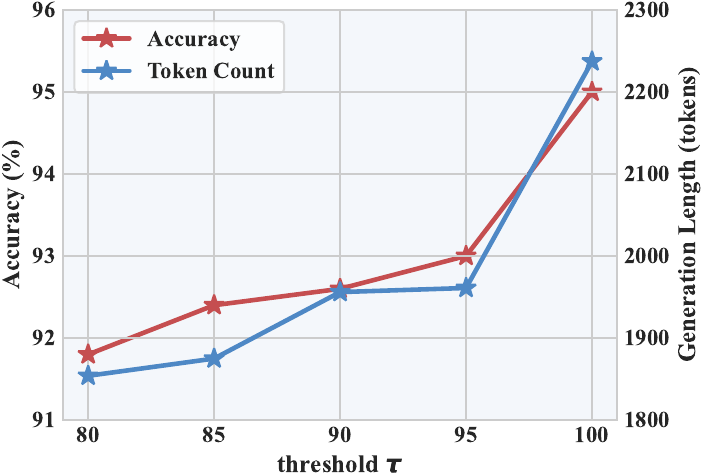}
    \caption{Impact of threshold $\tau$.}
    \label{fig:threshold}
\end{figure}

Figure~\ref{fig:threshold} presents the performance of DTSR on the MATH-500 dataset across different threshold values of $\tau$.
When the threshold is low, accuracy drops significantly because the model tends to terminate reasoning prematurely, before its thought process is fully developed.
This leads to incorrect answers, even though the overall generation length becomes somewhat shorter.
These results indicate that the optimal exit point for reasoning is achieved when $\tau = 100$, at which the model shows strong confidence in the sufficiency of its reasoning process.

\subsection{Analysis of Inference Latency}

\begin{table}[t]
\centering
\begin{tabular}{lcccc}
\toprule
\textbf{Method} 
& \multicolumn{2}{c}{\textbf{MATH-500}} 
& \multicolumn{2}{c}{\textbf{AMC}} \\
\cmidrule(lr){2-3} \cmidrule(lr){4-5}
& \textbf{Len} & \textbf{Latency}
& \textbf{Len} & \textbf{Latency}  \\
\midrule
Vanilla & 4480 & 3.1 &  6562 &  8.3\\
DEER    & 2601 & 4.2 &  4514 &  15.6\\
DTSR    & \textbf{2247} & \textbf{1.9} & \textbf{3975} & \textbf{6.2} \\
\bottomrule
\end{tabular}
\caption{Comparison of Inference Latency (s).}
\label{tab:len_latency}
\end{table}

Table~\ref{tab:len_latency} reports the end-to-end inference latency of DTSR on the MATH-500 and AMC datasets. The results show that our method not only substantially reduces the generation length, but also achieves a significant reduction in inference latency, yielding improvements of approximately 25\%–40\%.
In contrast, although DEER reduces the final generation length, it unexpectedly increases inference latency.
This performance gap can be attributed to two key factors. First, we enable the continuous batching strategy in vLLM, which allows different samples to interleave between the generation and evaluation phases. 
In contrast, DEER adopts a synchronous scheme that requires all samples to complete generation before a collective evaluation, resulting in idle computation and increased latency. 
Second, we introduce a token interval parameter $k$, which substantially reduces redundant checks, thereby further improving overall inference efficiency.

\begin{figure}[t]
    \centering
    \includegraphics[width=\columnwidth]{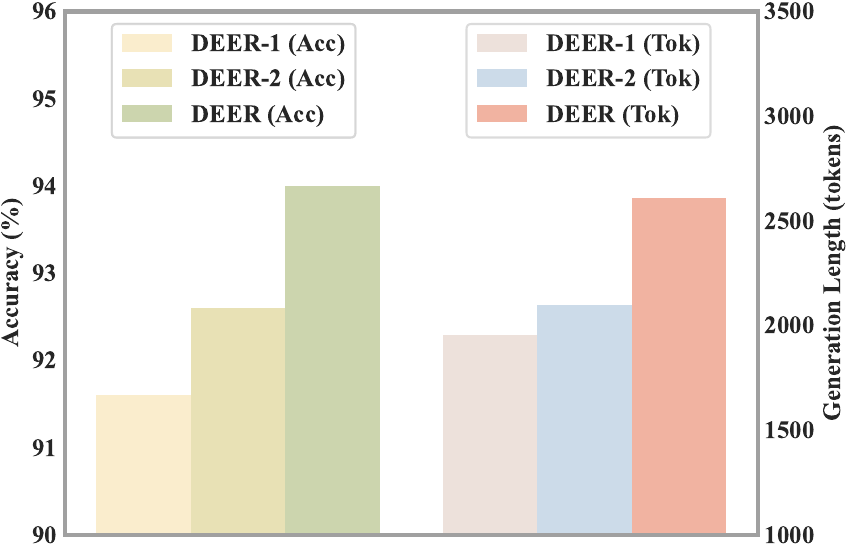}
    \caption{Comparison between DEER and its two variants, DEER-1 and DEER-2.}
    \label{fig:overconfidence}
\end{figure}

\subsection{Overconfidence in LRMs}

Here, we discuss the issue of overconfidence in LRM.
In fact, DEER has observed in experiments that the Qwen-3 series models tend to be overconfident in their confidence predictions.
In the original setup, DEER adds an early-exit condition on top of the confidence score derived from intermediate answers: the model is required to generate \texttt{</think>} after producing a trial answer.
To further investigate this issue, we derive two variants from DEER:
\begin{itemize}[leftmargin=*]
    \item \textbf{DEER-1:} The early-exit decision is based solely on the confidence score of the trial answer.
    \item \textbf{DEER-2:} Whether the model generates \texttt{</think>} after producing an intermediate answer is used as the criterion for early exit.
\end{itemize}
Figure~\ref{fig:overconfidence} presents the performance of DEER and its two variants, DEER-1 and DEER-2, on MATH-500.
DEER-1, which operates without the constraint of generating \texttt{</think>}, achieves the lowest accuracy, highlighting the model’s overconfidence in its generated answers.
This indicates that relying solely on the confidence score of a trial answer is insufficient for determining when to exit reasoning.
In contrast, DEER-2, which uses the simple criterion of whether \texttt{</think>} is generated, attains higher accuracy than DEER-1.
The generation of \texttt{</think>} after producing an answer is a behavior learned during training, serving as a signal of the model’s self-assessment of answer correctness.
These results suggest that leveraging the model’s intrinsic self-evaluation is a more reliable approach for guiding early-exit reasoning than using externally designed criteria.

\subsection{Self-Evaluation Paradigms}
\label{selfeval}
In DTSR, self-evaluation is conducted from a third-person perspective rather than by having the model directly score its own thoughts.
the two approaches are illustrated in Figure~\ref{fig:compare}.
To compare them, we introduce another variant, DTSR-1, in which an additional prompt is added during the intermediate reasoning process to ask the model to directly evaluate the adequacy of its prior reasoning.
The results on MATH-500 are shown in Table~\ref{table:self-evaluation}.
It is clear that DTSR-1 performs worse than DTSR in both accuracy and generation length. This is primarily because the model is unable to accurately assess the sufficiency of its reasoning during the process. As the saying goes, “the observer sees clearly what the participant cannot”.
Only by evaluating from a third-person perspective and separating reasoning generation from the judgment of its adequacy can the model demonstrate its full reasoning capability.

\begin{table}[t]
\centering
\normalsize
\begin{tabular}{lcc}
\toprule
\textbf{Method} & \textbf{Accuracy ($\uparrow$)} & \textbf{Length($\downarrow$)} \\
\midrule
DTSR & 95.0 & 2247 \\
DTSR-1 ($\tau$=100) & 94.2 & 3254 \\
DTSR-1 ($\tau$=95) & 93.8 & 2457 \\
\bottomrule
\end{tabular}
\caption{Comparison between DTSR and DTSR-1.}
\label{table:self-evaluation}
\end{table}

\subsection{Case Study}

We conduct case studies on three types of problems: mathematical, programming, and open-ended.
Figure~\ref{fig:case} shows the detailed reasoning process of Qwen3-14B on a math problem.
At the beginning, the model generates reasoning normally.
When it produces the reflection signal, it starts to evaluate the adequacy of its previous thoughts and gives a score of 75, which is below the early-exit threshold $\tau$.
As a result, the model returns to the reflection signal position to continue reasoning.
In the third round of reasoning, several reflection signals appear, but the model performs the thought sufficiency check only at the last one. This is because the number of generated tokens has not yet reached the token interval $k$, and redundant checks are skipped.
After several checks, the model finally gives a score of 100 on the fourth attempt, satisfying the early-exit condition. It then appends </think> and proceeds to make a conclusion, generating the final answer.
Figure~\ref{fig:case1} shows how DTSR solves a programming problem, and Figure~\ref{fig:case2} presents the comparison between Vanilla and DTSR on an open-ended question.
Across all cases, DTSR reduces token usage substantially while maintaining performance, improving reasoning efficiency and alleviating excessive thinking.

\section{Related work}
Recently, Large Reasoning Models (LRMs) have demonstrated remarkable performance on complex reasoning tasks by generating long Chains of Thought (CoT) for deep reasoning~\citep{DBLP:journals/corr/abs-2412-16720,Guo2025DeepSeek}.
However, recent studies~\citep{chen2025do,cuadron2025dangeroverthinkingexaminingreasoningaction} have revealed the issue of overthinking, where models continue to reason even after obtaining the correct answer, leading to redundant computational overhead.
To address this, prior research has explored a variety of approaches to enable efficient reasoning~\citep{feng2025efficientreasoningmodelssurvey}, which can be broadly categorized into three directions: (Input) prompt-guided methods, (Model) post-training methods, and (Output) decoding-intervention methods.

\paragraph{Prompt-guided Methods}
Such methods leverage carefully designed prompts to guide the model in performing efficient reasoning.
Concise CoT~\citep{Renze_2024} simply adds the instruction “Be concise” to the prompt, whereas TALE~\citep{han-etal-2025-token} first estimates an optimal token budget and then incorporates it into the prompt to encourage shorter reasoning chains.
Chain-of-Draft~\citep{xu2025chaindraftthinkingfaster} constrains the model to preserve only the most essential reasoning drafts at each step, thereby reducing token usage in few-shot settings.
Although easy to implement, these approaches~\citep{lee2025how} are limited in compressing long reasoning chains, and therefore better suited for use alongside other methods.

\paragraph{Post-training Methods}
These methods typically employ supervised fine-tuning (SFT) using variable-length CoT data~\citep{yu2024distilling,liu2024can,ma-etal-2025-cot}, or leverage reinforcement learning (RL) with length-based rewards~\citep{DBLP:journals/corr/abs-2501-12599,DBLP:journals/corr/abs-2502-04463,yeo2025demystifying,yi2025shorterbetterguidingreasoningmodels}.
L1~\citep{aggarwal2025l1controllinglongreasoning} and Elastic Reasoning~\citep{xu2025scalablechainthoughtselastic} incorporate length constraints into the reward function, training models to flexibly adjust the length of reasoning chains under different token budgets.
Recently, numerous studies~\citep{lou2025adacotparetooptimaladaptivechainofthought,jiang2025thinkneedlargehybridreasoning,fang2025thinklessllmlearnsthink} have focused on training hybrid reasoning models that adaptively choose between concise responses and long-chain reasoning based on task complexity.
While these methods attempt to address the problem at the fundamental level, their high training costs restrict most experiments to smaller models (e.g., 1.5B or 7B), limiting their applicability to larger models.

\paragraph{Decoding-intervention Methods}
Such methods intervene during decoding to adjust generation, guiding models to produce answers earlier~\citep{qiao2025conciseconfidenceguidedcompressionstepbystep,huang2025efficientreasoninglargereasoning} or exit reasoning sooner~\citep{liu2025answerconvergencesignalearly,jiang2025flashthinkearlyexitmethod,zhang2025reasoning}.
NoWAIT~\citep{wang2025waitdontneedwait} attempts to suppress explicit self-reflection by blocking tokens like Wait and Hmm during reasoning. However, the model can readily bypass this restriction by generating other tokens.
Dynasor-CoT~\citep{fu2025reasoning} probes the model’s intermediate answers at fixed token intervals and stops thinking if three consecutive answers are consistent, whereas DEER~\citep{yang2025dynamicearlyexitreasoning} tracks intermediate answers at reasoning transition points and determines whether to terminate reasoning based on their confidence scores.
However, both approaches rely on probing intermediate answers to determine stopping points, which works well only for problems with definite answers, but is less effective for long-form or open-ended outputs.
In contrast, our approach effectively overcomes this limitation.

\section{Conclusion}
In this work, we alleviate the inefficiency of LRMs by introducing Dynamic Thought Sufficiency in Reasoning (DTSR), a novel framework that enables models to dynamically determine when their CoT is sufficient to yield a reliable answer.
By integrating reflection signal monitoring with thought sufficiency checks, DTSR provides a principled mechanism for mitigating overthinking while maintaining strong reasoning performance. 
Experiments on Qwen3 models demonstrate that DTSR achieves substantial reductions in reasoning length with only minimal performance degradation, highlighting its effectiveness in improving efficiency.
We further discuss overconfidence in LRMs and the self-evaluation paradigm, shedding light on how models effectively assess the sufficiency of their CoT.
This exploration offers a fresh perspective and valuable insights for guiding early-exit reasoning.



\section*{Limitations}
While DTSR demonstrates promising results in improving reasoning efficiency, we identify some possible limitations as follows.
\begin{itemize}[leftmargin=*]
    \setlength{\itemsep}{0pt}
    \setlength{\parskip}{0pt}
    \item Due to computational resource constraints, we conduct experiments only on LLMs up to 32B.
    \item This work focuses exclusively on textual reasoning tasks, including mathematics and code. We have not yet extended our method to multimodal reasoning or agent scenarios, which we leave for future exploration.

\end{itemize}
 
\bibliography{custom}

\appendix
\section{Reflection Signal}
In this paper, we take the following keywords as signals for LRM reflection:
\begin{table}[h]
\centering
\begin{tabular}{l}
\hline
``Wait'', ``Alternative'', ``Alternatively''\\
``But wait'', ``But let me'', ``But let's''\\
\hline
\end{tabular}
\caption{Reflection Signal List}
\label{tab:Reflection_Signal}
\end{table}

\paragraph{Comparison with NoWAIT}
NoWAIT suppresses all reflection-related tokens, which can disrupt intrinsic reasoning and reduce accuracy. In contrast, DTSR preserves normal reasoning and triggers sufficiency evaluation only upon detecting reflection signals, achieving significant generation-length reduction without harming performance.

\section{Prompt Template}
Figure~\ref{fig:template} presents the prompt template used for thought sufficiency check stage.
Figure~\ref{fig:compare} illustrates the differences between the prompt templates of DTSR and its variant, DTSR-1.

\section{More Experiments}
\paragraph{Comparison with Training-based Mehtods}
We further compare DTSR with more baselines. For training-based methods, we include RL + Length Penalty~\citep{arora2025training} and S-GRPO~\citep{dai2025sgrpo}.
RL + Length Penalty penalizes overly long correct responses by assigning rewards based on their deviation from the mean length.
S-GRPO uses rule-based outcome rewards to regulate intermediate reasoning, encouraging LLMs to generate high-quality thoughts earlier and to exit once sufficient reasoning is reached, thereby improving efficiency while maintaining accuracy.
As shown in Table~\ref{table:training_based}, DTSR achieves a generation length comparable to training-based methods. Although it exhibits slightly lower accuracy due to the absence of additional training, DTSR remains competitive while requiring no extra training.

\paragraph{Comparison with Training-free Mehtods}
We also compare DTSR with two recent efficient reasoning methods, TALE~\citep{han-etal-2025-token} and Chain-of-Draft (CoD)~\citep{xu2025chaindraftthinkingfaster}.
As shown in Table~\ref{table:tale_cod}, TALE achieves almost no reduction in generation length, while CoD substantially reduces token usage at the cost of a significant accuracy drop. In contrast, DTSR strikes a better balance between accuracy and generation length.

\paragraph{Performance under Different Decoding Strategies}
In the main experiments, we adopt a decoding configuration with temperature set to 0.6 and top\_p set to 0.95. Table~\ref{table:greedy} reports the results under greedy decoding. The results show that DTSR maintains strong performance across different decoding strategies, further demonstrating the robustness and applicability of our method.

\section{Case Study}
\begin{figure*}[t]
	\centering
    \includegraphics[width=\textwidth]{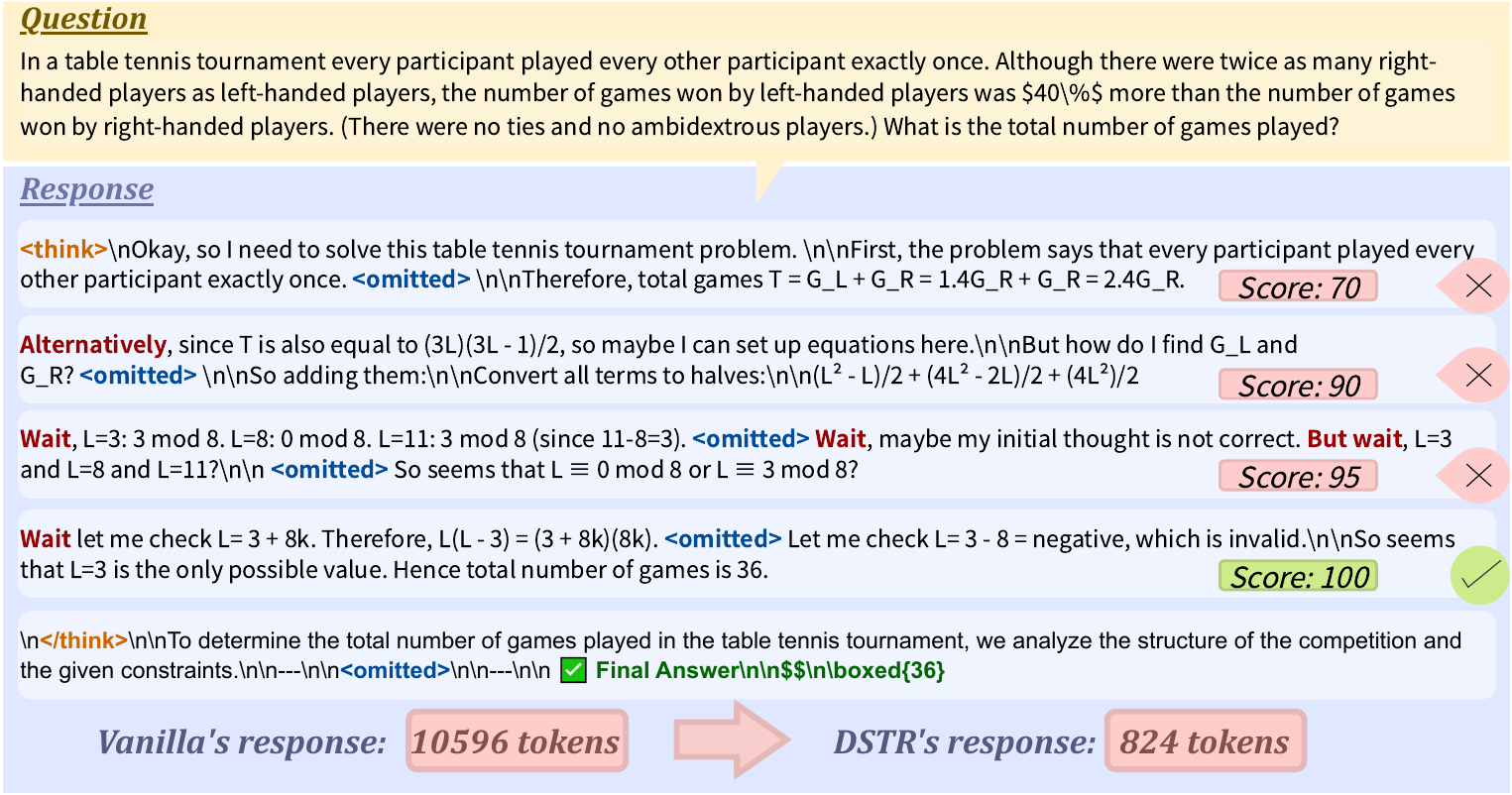}
    \caption{An example illustrating the reasoning process of DTSR.}
  \label{fig:case}
\end{figure*}

\begin{figure*}[t]
	\centering
    \includegraphics[width=\textwidth]{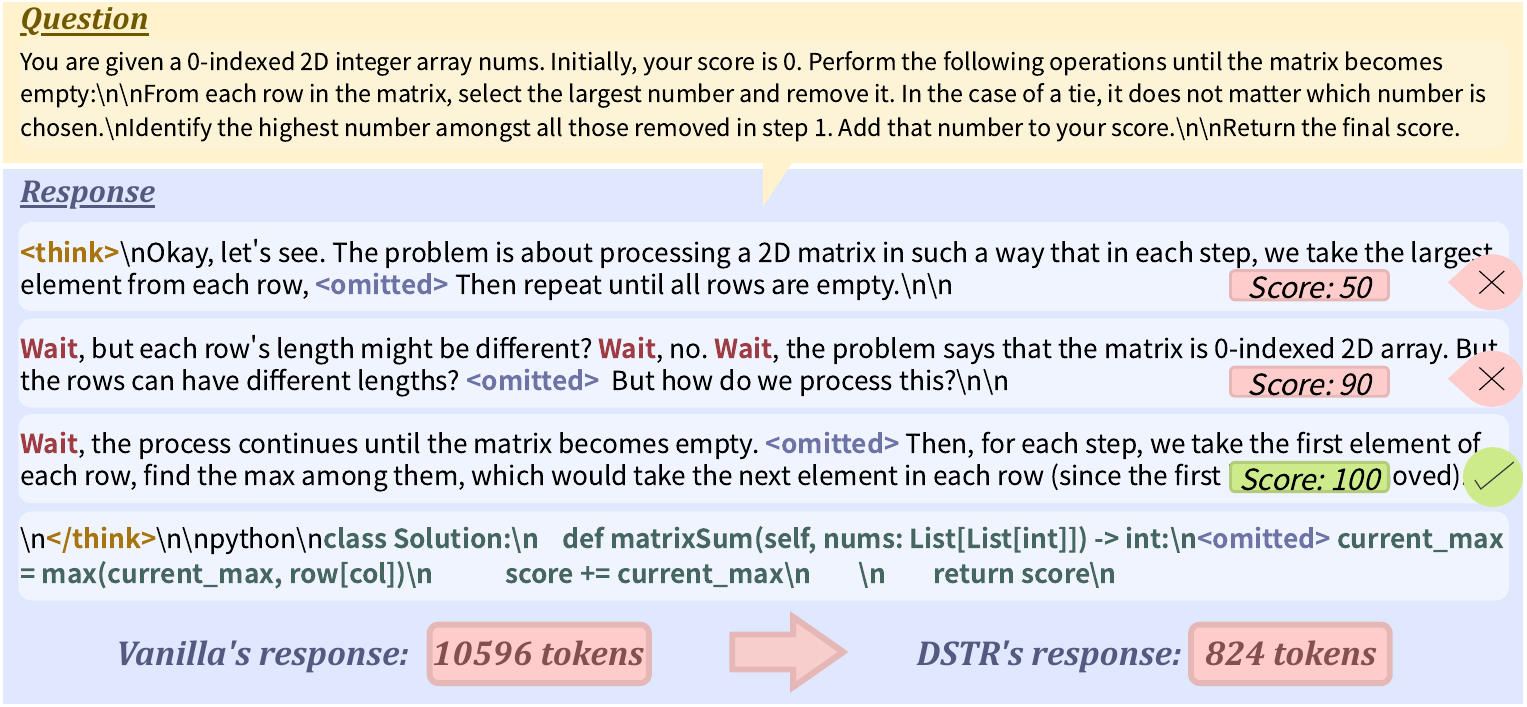}
    \caption{An example illustrating the reasoning process of DTSR on a promgramming problem.}
  \label{fig:case1}
\end{figure*}

Figures~\ref{fig:case},~\ref{fig:case1} and ~\ref{fig:case2} present representative examples of Qwen-14B solving mathematical, programming, and open-ended problems under the DTSR framework, respectively.

\begin{figure*}[t]
	\centering
    \includegraphics[width=\textwidth]{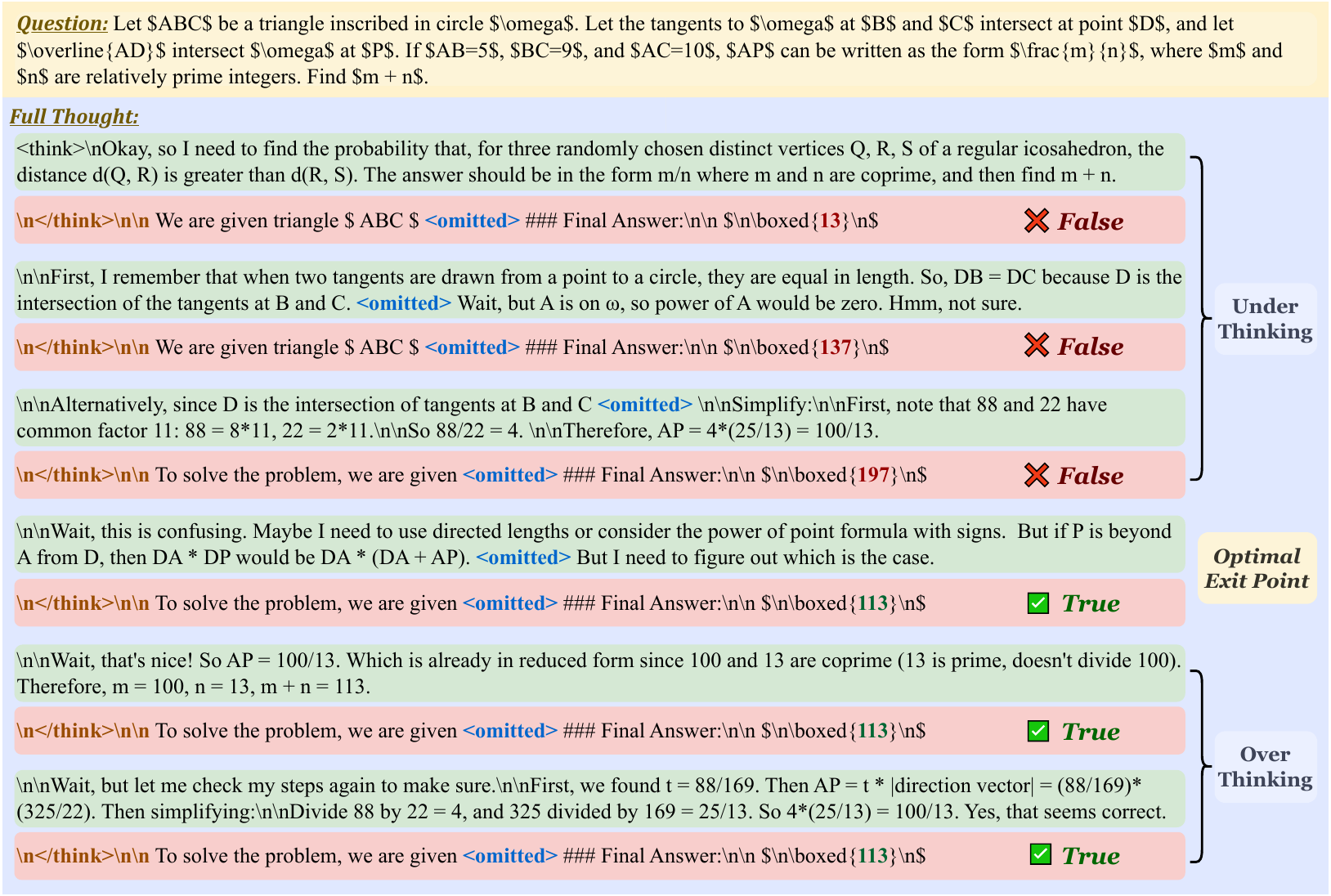}
    \caption{An illustrative example showing how we identify the optimal exit point. After each generated sentence, we append \texttt{</think>} to force the model to produce an answer. The correctness of the resulting answer is then used to determine whether the current reasoning state is sufficient and corresponds to the optimal exit point.}
  \label{fig:example}
\end{figure*}

\begin{figure*}[t] 
\centering
\begin{tcolorbox}[colback=white, colframe=white!60!black, coltitle=black, colbacktitle=brown!40!white, 
title=Thought Sufficiency Check Prompt Template,
fonttitle=\bfseries, enhanced, 
attach boxed title to top center={yshift=-2mm},
width=\textwidth] 
\begin{verbatim}
<|im_start|>users
Assess the confidence that the provided thought is sufficient to solve the question.
Output only in the format: Confidence: <number>, where <number> is a real value
between 0 and 100. Here, 100 means the thought is fully sufficient to reach the 
final answer directly without additional reasoning, and 0 means the thought is 
completely insufficient to do so.

### Question
{question}

### Thought
{thought}
<|im_end|>
<|im_start|>assistant
<think>

</think>

Confidence:
\end{verbatim}
\end{tcolorbox}
\caption{Thought Sufficiency Check Prompt Template}
\label{fig:template}
\end{figure*}

\begin{figure*}[t] 
\centering
\begin{tcolorbox}[colback=white, colframe=white!60!black, coltitle=black, colbacktitle=brown!40!white, 
title=DTSR,
fonttitle=\bfseries, enhanced, 
attach boxed title to top center={yshift=-2mm},
width=\textwidth] 
\begin{verbatim}
<|im_start|>users
Assess the confidence that the provided thought is sufficient to solve the question.
Output only in the format: Confidence: <number>, where <number> is a real value
between 0 and 100. Here, 100 means the thought is fully sufficient to reach the 
final answer directly without additional reasoning, and 0 means the thought is 
completely insufficient to do so.

### Question
{How many positive whole-number divisors does 196 have?}

### Thought
{<think>\nOkay, so I need to figure out how many positive whole-number divisors the
number 196 has. Hmm, let me think. I remember that to find the number of divisors
of a number, you first have to find its prime factorization. Once you have that,
you can use exponents to determine the total number of divisors. Let me try to
recall exactly how that works.}
<|im_end|>
<|im_start|>assistant
<think>

</think>

Confidence:
\end{verbatim}
\end{tcolorbox}


\begin{tcolorbox}[colback=white, colframe=white!60!black, coltitle=black, colbacktitle=brown!40!white, 
title=DTSR-1,
fonttitle=\bfseries, enhanced, 
attach boxed title to top center={yshift=-2mm},
width=\textwidth] 
\begin{verbatim}
<|im_start|>system
Please reason step by step, and put your final answer within \\boxed{}.
<|im_end|>
<|im_start|>users
How many positive whole-number divisors does 196 have?
<|im_end|>
<|im_start|>assistant
<think>\nOkay, so I need to figure out how many positive whole-number divisors the
number 196 has. Hmm, let me think. I remember that to find the number of divisors
of a number, you first have to find its prime factorization. Once you have that,
you can use exponents to determine the total number of divisors. Let me try to
recall exactly how that works.

Assess the confidence that the provided thought is sufficient to solve the question.
Output only in the format: Confidence: <number>, where <number> is a real value
between 0 and 100. Here, 100 means the thought is fully sufficient to reach the 
final answer directly without additional reasoning, and 0 means the thought is 
completely insufficient to do so.

Confidence:
\end{verbatim}
\end{tcolorbox}

\caption{Differences between DTSR and its variant DTSR-1.}
\label{fig:compare}
\end{figure*}

\clearpage

\begin{table*}[t]
\centering
\small
\label{table1}
\renewcommand{\arraystretch}{1}
\begin{tabular}{@{}lccccccccccll@{}} 
\toprule
 \multirow{2}{*}{\textbf{Method}} 
 & \multicolumn{2}{c}{\textbf{GSM8K}} & \multicolumn{2}{c}{\textbf{MATH-500}} & \multicolumn{2}{c}{\textbf{AMC}} & \multicolumn{2}{c}{\textbf{GPQA}} & \multicolumn{2}{|c}{\textbf{Overall}} \\
    & {Acc$\uparrow$} & {Tok$\downarrow$} & {Acc$\uparrow$} & {Tok$\downarrow$} & {Acc$\uparrow$} & {Tok$\downarrow$} & {Acc$\uparrow$} & {Tok$\downarrow$} & {Acc$\uparrow$} & {Tok$\downarrow$}\\ 
\midrule
\multicolumn{11}{c}{\textit{\textbf{Qwen3-14B}}} \\
\midrule
\textit{Vanilla} & 96.2 & 1672 & 95.4 & 4503 & 97.5 & 6562 & 65.7 & 7513 & \multicolumn{1}{|l}{88.7} & 5062 \\
\textit{RL + Length Penalty} & 95.8 & 1090 & 95.8 & 2866 & 97.5 & 5059 & 65.4 & 4949 & \multicolumn{1}{|l}{88.6} & 3491 \\
\textit{S-GRPO} & 96.3 & 952 & 96.4 & 2652 & 100 & 4538 & 67.6 & 4537 & \multicolumn{1}{|l}{90.1} & 3170 \\
\textit{DTSR} & 96.2 & 849 & 95.0 & 2247 & 97.5 & 3975 & 66.2 & 5916 & \multicolumn{1}{|l}{88.7} & 3246 \\
\bottomrule
\end{tabular}
\caption{Comparison of DTSR with training-based efficient reasoning methods. "Acc" denotes accuracy, and "Tok" denotes token count. $\uparrow$ indicates that higher values are better, while $\downarrow$ indicates that lower values are better.}
\label{table:training_based}
\end{table*}

\begin{table*}[t]
\centering
\small
\label{table1}
\renewcommand{\arraystretch}{1}
\begin{tabular}{@{}lccccccccccll@{}} 
\toprule
 \multirow{2}{*}{\textbf{Method}} 
 & \multicolumn{2}{c}{\textbf{GSM8K}} & \multicolumn{2}{c}{\textbf{MATH-500}} & \multicolumn{2}{c}{\textbf{AMC}} & \multicolumn{2}{c}{\textbf{GPQA}} & \multicolumn{2}{|c}{\textbf{Overall}} \\
    & {Acc$\uparrow$} & {Tok$\downarrow$} & {Acc$\uparrow$} & {Tok$\downarrow$} & {Acc$\uparrow$} & {Tok$\downarrow$} & {Acc$\uparrow$} & {Tok$\downarrow$} & {Acc$\uparrow$} & {Tok$\downarrow$}\\ 
\midrule
\multicolumn{11}{c}{\textit{\textbf{Qwen3-14B}}} \\
\midrule
\textit{Vanilla} & 96.2 & 1672 & 95.4 & 4503 & 97.5 & 6562 & 65.7 & 7513 & \multicolumn{1}{|l}{88.7} & 5062 \\
\textit{TALE} & 95.7 & 1241 & 94.6 & 4484 & 95.0 & 6623 & 65.6 & 7138 & \multicolumn{1}{|l}{87.7} & 4872 \\
\textit{CoD} & 85.7 & 648 & 75.2 & 2359 & 72.5 & 4122 & 51.0 & 1177 & \multicolumn{1}{|l}{71.1} & 2077 \\
\textit{DTSR} & 96.2 & 849 & 95.0 & 2247 & 97.5 & 3975 & 66.2 & 5916 & \multicolumn{1}{|l}{\textbf{88.7}} & \textbf{3246} \\
\bottomrule
\end{tabular}
\caption{Comparison of DTSR with TALE and Chain-of-Draft (CoD). "Acc" denotes accuracy, and "Tok" denotes token count. $\uparrow$ indicates that higher values are better, while $\downarrow$ indicates that lower values are better.}
\label{table:tale_cod}
\end{table*}

\begin{table*}[!t]
\centering
\small
\label{table1}
\renewcommand{\arraystretch}{1}
\begin{tabular}{@{}lccccccccccll@{}} 
\toprule
 \multirow{2}{*}{\textbf{Method}} 
 & \multicolumn{2}{c}{\textbf{GSM8K}} & \multicolumn{2}{c}{\textbf{MATH-500}} & \multicolumn{2}{c}{\textbf{AMC}} & \multicolumn{2}{c}{\textbf{GPQA}} & \multicolumn{2}{c}{\textbf{Olympiad}} & \multicolumn{2}{|c}{\textbf{Overall}} \\
    & {Acc$\uparrow$} & {Tok$\downarrow$} & {Acc$\uparrow$} & {Tok$\downarrow$} & {Acc$\uparrow$} & {Tok$\downarrow$} & {Acc$\uparrow$} & {Tok$\downarrow$} & {Acc$\uparrow$} & {Tok$\downarrow$} & {Acc$\uparrow$} & {Tok$\downarrow$} \\ 
\midrule
\multicolumn{13}{c}{\textit{\textbf{Qwen3-14B}}} \\
\midrule
\textit{Vanilla} & 96.3 & 1676 & 94.8 & 4511 & 97.5 & 6998 & 64.1 & 7382 & 65.6 & 8644 & \multicolumn{1}{|l}{83.7} & 5842\\
\textit{DTSR} & 96.1 & 814 & 94.6 & 2296 & 97.5 & 3977 & 65.6 & 6103 & 66.8 & 5967 & \multicolumn{1}{|l}{{\textbf{84.1}}} & 
\textbf{3831} \\
\bottomrule
\end{tabular}
\caption{Experimental results under greedy decoding. "Acc" denotes accuracy, and "Tok" denotes token count. $\uparrow$ indicates that higher values are better, while $\downarrow$ indicates that lower values are better.}
\label{table:greedy}
\end{table*}

\begin{figure*}[t]
\centering
\begin{tcolorbox}[colback=gray!5!white,
  colframe=black!75!black,
  fonttitle=\bfseries,
  colbacktitle=gray!85!black,
  title=Case Example: Open Question,
  width=\textwidth,
  boxrule=0.3pt,
  arc=2pt,
  left=5pt, right=5pt, top=4pt, bottom=4pt,
  before skip=8pt, after skip=8pt,
  breakable,
  enhanced jigsaw
]

\small
\textbf{Question:}\\
Suppose gravity on Earth suddenly became half as strong.\\
How would human architecture, biology, and transportation adapt over time?\\
\rule{\linewidth}{0.4pt}
\textbf{Vanilla:}\\
<think>...</think>\\\\If Earth's gravity suddenly became half as strong (i.e., gravitational acceleration reduced from 9.8 m/s² to 4.9 m/s²), the effects on human architecture, biology, and transportation would be profound and multifaceted, with immediate and long-term adaptations across these domains.\\
1. Architecture\\- Reduced Structural Load: Buildings, bridges, and other infrastructure would require significantly less material and reinforcement. For example, skyscrapers could be taller or constructed with lighter materials (e.g., advanced composites instead of steel) due to decreased compressive forces.\\- Design Adjustments: Foundations might be smaller or less deep, and load-bearing walls could be thinner. However, engineers would need to account for new challenges, such as wind forces (which are less mitigated by gravity) and seismic activity (potentially altered by changes in Earth's mass or tectonic dynamics).\\- Fluid Dynamics: Plumbing systems might require adjustments to handle reduced pressure in water distribution, and drainage systems could become less efficient due to slower fluid flow under lower gravity...\\2. Human Biology\\- Reduced Physical Strain: Humans would feel lighter, making movement easier and reducing joint stress. People might jump higher, run faster, and experience less fatigue in daily activities.\\- Musculoskeletal Changes: Over time, muscle atrophy and bone density loss could occur due to reduced mechanical loading. This would mirror the effects of prolonged microgravity on astronauts, potentially leading to osteoporosis and weakened ligaments.\\- Cardiovascular Adaptation: The heart would pump less forcefully, and blood pressure might decrease. Circulatory systems might adjust to reduced gravitational \"pull,\" potentially altering the distribution of fluids in the body...\\3. Transportation\\- Easier Space Travel: Launching spacecraft would require less fuel to escape Earth's gravity, making interplanetary missions more feasible and cost-effective.\\- Vehicle Design: Cars and other vehicles might become lighter and more agile, with reduced traction requirements...\\
\rule{\linewidth}{0.4pt}
\textbf{DTSR:}\\
<think>...</think>\\
If Earth's gravity suddenly became half as strong, the following adaptations would occur in architecture, biology, and transportation over time:\\
1. Architecture\\- Structural Design: Buildings would require less material and reinforcement due to reduced gravitational load. Foundations could be shallower, and skyscrapers might reach greater heights since the structural stress is halved.\\- Material Use: Lighter materials (e.g., composites, expanded metals) might replace heavier ones like concrete and steel. Open spaces, larger windows, and more cantilevers could become common.\\- Wind Considerations: While gravity is lower, wind forces on tall structures might remain a limiting factor, potentially leading to more aerodynamic designs.\\
2. Biology\\- Musculoskeletal System: Humans would likely experience reduced bone density and muscle mass due to less gravitational stress. Over generations, evolutionary adaptations might include taller, more slender bodies with less robust bones and muscles.\\- Cardiovascular System: The heart might become smaller, as it would not need to work as hard to pump blood against gravity. However, if atmospheric pressure decreases (due to lower gravity), respiratory adaptations (e.g., larger lungs, higher hemoglobin levels) could emerge to compensate for lower oxygen availability.\\- Fluid Distribution: Fluids (e.g., blood, interstitial fluid) would pool more in the upper body, leading to facial puffiness and potential changes in circulatory dynamics.\\
3. Transportation\\- Ground Vehicles: Vehicles would weigh less, improving fuel efficiency and acceleration. Braking systems might need redesigning due to reduced friction (normal force is halved). Heavier cargo could be transported with the same infrastructure.\\- Aircraft: If atmospheric pressure remains constant, aircraft would require less lift, enabling smaller, more efficient designs. However, if the atmosphere thins (due to lower gravity), air density would decrease, complicating lift generation. This might necessitate larger wings or more powerful engines.\\- Space Travel: Launching to space would require less energy, making interplanetary travel more feasible..
\\

\end{tcolorbox}
\caption{Comparison of Vanilla's and DTSR's answer on an open question.}
\label{fig:case2}
\end{figure*}

\end{document}